\documentclass[11pt,letterpaper]{article}
\usepackage{emnlp2017}
\usepackage{times}
\usepackage{latexsym}
\usepackage{enumitem}
\usepackage{svg}
\usepackage{tabularx}
\usepackage{hyperref}
\usepackage{multirow}
\usepackage[super]{nth}
\usepackage{array}

\newcolumntype{L}{>{\centering\arraybackslash}m{3cm}}

\emnlpfinalcopy

\def\emnlppaperid{***}

\title{Seernet at EmoInt-2017: Tweet Emotion Intensity Estimator}

\author{Venkatesh Duppada \and Sushant Hiray \\
  Seernet Technologies, LLC \\ \{venkatesh.duppada, sushant.hiray\}@seernet.io }

\date{}

\begin{document}

\maketitle

\begin{abstract}
	The paper describes experiments on estimating emotion intensity in tweets using a generalized regressor system. The system combines lexical, syntactic and pre-trained word embedding
features, trains them on general regressors and finally combines the best performing models to create an ensemble. The proposed system stood \nth{3} out of 22 systems in the leaderboard of WASSA-2017 Shared Task on Emotion Intensity.
\end{abstract}

\section{Introduction}
Twitter, a micro-blogging and social networking site has emerged as a platform where people express themselves and react to events in real-time. It is estimated that nearly 500 million tweets are sent per day \footnote{\url{https://en.wikipedia.org/wiki/Twitter}}. Twitter data is particularly interesting because of its peculiar nature where people convey messages in short sentences using hashtags, emoticons, emojis etc. In addition, each tweet has meta data like location and language used by the sender. It's challenging to analyze this data because the tweets might not be grammatically correct and the users tend to use informal and slang words all the time. Hence, this poses an interesting problem for NLP researchers. Any advances in using this abundant and diverse data can help understand and analyze information about a person, an event, a product, an organization or a country as a whole. Many notable use cases of the twitter can be found here\footnote{\url{https://en.wikipedia.org/wiki/Twitter_usage}}.

Along the similar lines, \textbf{The Task 1 of WASSA-2017} \cite{wassatask2017} poses a problem of finding emotion intensity of four emotions namely anger, fear, joy, sadness from tweets. In this paper, we describe our approach and experiments to solve this problem. The rest of the paper is laid out as follows: Section 2 describes the system architecture, Section 3 reports results and inference from different experiments, while Section 4 points to ways that the problem can be further explored.

\section{System Description}


\begin{figure}
    \centering 
    \includegraphics[width=\columnwidth]{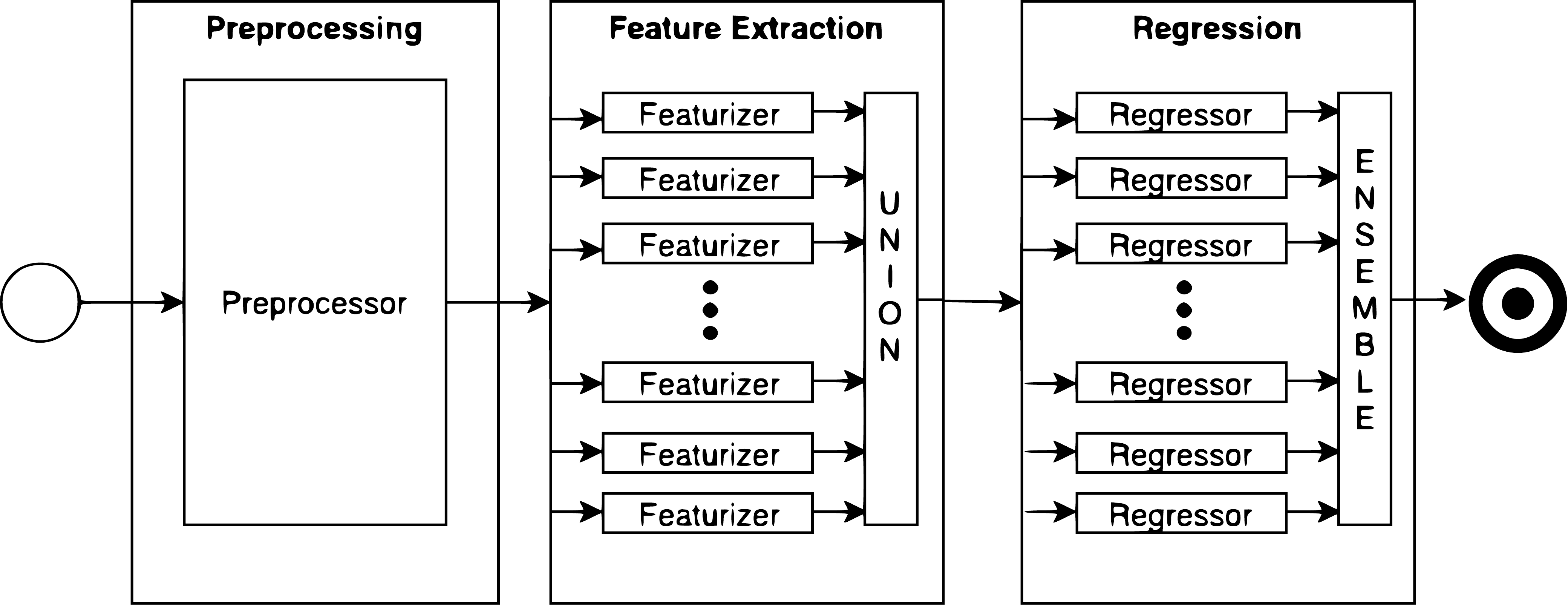}
	\caption{System Architecture}
	\label{fig:system_description}
\end{figure}

\subsection{Preprocessing}
The preprocessing step modifies the raw tweets before they are passed to feature extraction.
Tweets are processed using \textbf{tweetokenize} tool\footnote{\url{https://www.github.com/jaredks/tweetokenize}}. Twitter specific features are replaced as follows: username handles to {\tt USERNAME}, phone numbers to {\tt PHONENUMBER}, numbers to {\tt NUMBER}, URLs to {\tt URL} and times to {\tt TIME}. A continuous sequence of emojis is broken into individual tokens. Finally, all tokens are converted to lowercase.

\subsection{Feature Extraction} \label{feature}
Many tasks related to sentiment or emotion analysis depend upon affect, opinion, sentiment, sense and emotion lexicons. These lexicons associate words to corresponding sentiment or emotion metrics. On the other hand, the semantic meaning of words, sentences, and documents are preserved and compactly represented using low dimensional vectors \cite{mikolov2013distributed} instead of one hot encoding vectors which are sparse and high dimensional. Finally, there are traditional NLP features like word N-grams, character N-grams, Part-Of-Speech N-grams and word clusters which are known to perform well on various tasks.

Based on these observations, the feature extraction step is implemented as a union of different independent feature extractors (featurizers) in a light-weight and easy to use Python program EmoInt \footnote{To enable replicability, the code is open sourced at \url{https://github.com/SEERNET/EmoInt}.}. It comprises of all features available in the baseline model \cite{affectivetweets} \footnote{\url{https://www.github.com/felipebravom/AffectiveTweets}} along with additional feature extractors and bi-gram support. Fourteen such feature extractors have been implemented which can be clubbed into 3 major categories:

\begin{itemize}[noitemsep]
	\item Lexicon Features
	\item Word Vectors
	\item Syntax Features
\end{itemize}

\textbf{Lexicon Features}: AFINN \cite{nielsen2011new} word list are manually rated for valence with an integer between -5 (Negative Sentiment) and +5 (Positive Sentiment). Bing Liu \cite{hu2004mining} opinion lexicon extract opinion on customer reviews. +/-EffectWordNet \cite{choi2014+} by MPQA group are sense level lexicons. The NRC Affect Intensity \cite{mohammad2017word} lexicons provide real valued affect intensity. NRC Word-Emotion Association Lexicon \cite{mohammad2010emotions} contains 8 sense level associations (anger, fear, anticipation, trust, surprise, sadness, joy, and disgust) and 2 sentiment level associations (negative and positive). Expanded NRC Word-Emotion Association Lexicon \cite{bravo2016determining} expands the NRC word-emotion association lexicon for twitter specific language. NRC Hashtag Emotion Lexicon \cite{mohammad2015using} contains emotion word associations computed on emotion labeled twitter corpus via Hashtags. NRC Hashtag Sentiment Lexicon and Sentiment140 Lexicon \cite{mohammad2013nrc} contains sentiment word associations computed on twitter corpus via Hashtags and Emoticons. SentiWordNet \cite{baccianella2010sentiwordnet} assigns to each synset of WordNet three sentiment scores: positivity, negativity, objectivity. Negation lexicons collections are used to count the total occurrence of negative words. In addition to these, SentiStrength \cite{thelwall2010sentiment} application which estimates the strength of positive and negative sentiment from tweets is also added. 

\textbf{Word Vectors}: We focus primarily on the word vector representations (word embeddings) created specifically using the twitter dataset. GloVe \cite{pennington2014glove} is an unsupervised learning algorithm for obtaining vector representations for words. 200-dimensional GloVe embeddings trained on 2 Billion tweets are integrated. Edinburgh embeddings \cite{bravo2015unlabelled} are obtained by training skip-gram model on  Edinburgh corpus \cite{petrovic2010edinburgh}. Since tweets are abundant with emojis, Emoji embeddings \cite{eisner2016emoji2vec} which are learned from the emoji descriptions have been used. Embeddings for each tweet are obtained by summing up individual word vectors and then dividing by the number of tokens in the tweet.

\textbf{Syntactic Features}: Syntax specific features such as Word N-grams, Part-Of-Speech N-grams \cite{owoputi2013improved}, Brown Cluster N-grams \cite{brown1992class} obtained using TweetNLP \footnote{\url{http://www.cs.cmu.edu/~ark/TweetNLP/}} project have been integrated into the system. 

The final feature vector is the concatenation of all the individual features. For example, we concatenate average word vectors, sum of NRC Affect Intensities, number of positive and negative Bing Liu lexicons, number of negation words and so on to get final feature vector. The scaling of final features is not required when used with gradient boosted trees. However, scaling steps like standard scaling (zero mean and unit normal) may be beneficial for neural networks as the optimizers work well when the data is centered around origin.

A total of fourteen different feature extractors have been implemented, all of which can be enabled or disabled individually to extract features from a given tweet.

\subsection{Regression}
The dev data set \cite{MohammadB17starsem} in the competition was small hence, the train and dev sets were merged to perform 10-fold cross validation. On each fold, a model was trained and the predictions were collected on the remaining dataset. The predictions are averaged across all the folds to generalize the solution and prevent over-fitting. As described in Section \ref{feature}, different combinations of feature extractors were used. After performing feature extraction, the data was then passed to various regressors Support Vector Regression, AdaBoost, RandomForestRegressor, and, BaggingRegressor of sklearn \cite{scikit-learn}. Finally, the chosen top performing models had the least error on evaluation metrics namely Pearson's Correlation Coefficient and Spearman's rank-order correlation.

\subsection{Parameter Optimization} \label{optimization}
In order to find the optimal parameter values for the EmoInt system, an extensive grid search was performed through the scikit-Learn framework
over all subsets of the training set (shuffled), using stratified 10-fold cross validation and optimizing the Pearson's Correlation score. 
Best cross-validation results were obtained using AdaBoost meta regressor with base regressor as XGBoost \cite{chen2016xgboost} with 1000 estimators and 0.1 learning rate. Experiments and analysis of results are presented in the next section.

\section{Results and Analysis}


\subsection{Experimental Results}
As described in Section \ref{feature} various syntax features were used namely, Part-of-Speech tags, brown clusters of TweetNLP project. However, these didn't perform well in cross validation. Hence, they were dropped from the final system. While performing grid-search as mentioned in Section \ref{optimization}, keeping all the lexicon based features same, choice of combination of emoji vector and word vectors are varied to minimize cross validation metric. Table \ref{results} describes the results for experiments conducted with different combinations of word vectors. Emoji embeddings \cite{eisner2016emoji2vec} give better results than using plain GloVe and Edinburgh embeddings. Edinburgh embeddings outperform GloVe embeddings in \textbf{Joy} and \textbf{Sadness} category but lag behind in \textbf{Anger} and \textbf{Fear} category. The official submission comprised of the top-performing model for each emotion category. This system ranked \nth{3} for the entire test dataset and \nth{2} for the subset of the test data formed by taking every instance with a gold emotion intensity score greater than or equal to 0.5. Post competition, experiments were performed on ensembling diverse models for improving the accuracy. An ensemble obtained by averaging the results of the top 2 performing models outperforms all the individual models.

\begin{table*}[!htbp]
{
	\begin{tabular}{||c|c|c|c|c|c||}
		\hline
		\textbf{Emotion} & \textbf{Systems} & \textbf{Pearsonr}  & \textbf{Spearmanr} & \textbf{Pearsonr} $\geq$ 0.5 & \textbf{Spearmanr} $\geq$ 0.5 \\
		\hline\hline
				    
		\textbf{Anger}   & Baseline         & 0.639583           & 0.628180           & 0.510361                     & 0.475215                      \\
		                 & Em0-Ed1-Gl0      & 0.659566           & 0.628835           & 0.536701                     & 0.508762                      \\ 
		                 & Em1-Ed1-Gl0      & 0.660568           & 0.631893           & 0.536244                     & 0.511621                      \\
		                 & Em0-Ed0-Gl1\textbf{*}      & 0.675864 & 0.656034           & 0.529404                     & 0.512774                      \\
		                 & Em1-Ed0-Gl1      & 0.678214           & \textbf{0.658605}           & 0.527375                     & 0.510436                      \\
		                 & Ensemble         & \textbf{0.678477}  & 0.653964           & \textbf{0.540919}                     & \textbf{0.518851}                      \\
		\hline
				    
		\textbf{Fear}    & Baseline         & 0.631139           & 0.622047           & 0.476480                     & 0.432407                      \\
		                 & Em0-Ed1-Gl0      & 0.689571           & 0.66237            & 0.539250                     & 0.499864                      \\ 
		                 & Em1-Ed1-Gl0      & 0.695443           & 0.670438           & 0.542909                     & 0.500896                      \\
		                 & Em0-Ed0-Gl1      & 0.691143           & 0.667255           & 0.546867                     & 0.510041                      \\
		                 & Em1-Ed0-Gl1\textbf{*}      & 0.697630 & 0.676379           & 0.551465                     & 0.510265                      \\
		                 & Ensemble         & \textbf{0.705260}  & \textbf{0.683536}           & 0.\textbf{55641}                      & \textbf{0.513398}                      \\
		\hline
				    
		\textbf{Joy}     & Baseline         & 0.645597           & 0.652505           & 0.370499                     & 0.363184                      \\
		                 & Em0-Ed1-Gl0      & 0.696448           & 0.66237            & 0.539250                     & 0.499864                      \\ 
		                 & Em1-Ed1-Gl0      & 0.722115           & 0.720437           & 0.519821                     & 0.508484                      \\
		                 & Em0-Ed0-Gl1      & 0.689692           & 0.689883           & 0.472973                     & 0.470260                      \\
		                 & Em1-Ed0-Gl1\textbf{*}      & 0.714850 & 0.713558           & \textbf{0.551191}                     & \textbf{0.543565}                      \\
		                 & Ensemble         & \textbf{0.728093}  & \textbf{0.727970}           & 0.547213                     & 0.537690                      \\
		\hline
				    
		\textbf{Sadness} & Baseline         & 0.711998           & 0.711745           & 0.479049                     & 0.452047                      \\
		                 & Em0-Ed1-Gl0      & 0.737805           & 0.733999           & 0.547871                     & 0.524843                      \\ 
		                 & Em1-Ed1-Gl0\textbf{*}      & 0.744550 & 0.740893           & \textbf{0.554723}                     & 0.533571                      \\
		                 & Em0-Ed0-Gl1      & 0.731436           & 0.724570           & 0.542910                     & 0.536228                      \\
		                 & Em1-Ed0-Gl1      & 0.736081           & 0.731050           & 0.553460                     & \textbf{0.548944}                      \\
		                 & Ensemble         & \textbf{0.748901}  & \textbf{0.743589}           & 0.547213                     & 0.537690                      \\
		\hline \hline
				    
		\textbf{Average} & Baseline         & 0.657079           & 0.653619           & 0.479049                     & 0.452047                      \\
		                 & Em0-Ed1-Gl0      & 0.695847           & 0.680207           & 0.51998                      & 0.493755                      \\ 
		                 & Em1-Ed1-Gl0      & 0.705669           & 0.690915           & 0.538424                     & 0.513643                      \\
		                 & Em0-Ed0-Gl1      & 0.69703            & 0.684436           & 0.523038                     & 0.507326                      \\
		                 & Em1-Ed0-Gl1      & 0.706694           & 0.694898           & 0.545873                     & 0.528303                      \\
		                 & Official\textbf{*}         & 0.708267 & 0.696801           & 0.546913                     & 0.526018                      \\
		                 & Ensemble         & \textbf{0.715183}  & \textbf{0.702265}           & \textbf{0.55209}                      & \textbf{0.530501}                      \\
		\hline
				    
	\end{tabular}}
	\caption{\label{results} Evaluation Metrics for various systems. Systems are abbreviated as following: For example \texttt{Em1-Ed0-Gl1} implies Emoji embeddings and GloVe embeddings are included, Edinburgh embeddings are not included in features keeping other features same. Results marked with \textbf{*} corresponds to official submission. Results in \textbf{bold} are the best results corresponding to that metric.}
\end{table*}

\subsection{Feature Importance}
\begin{figure*}[!htbp]
	\center
	\includegraphics[width=1.0\textwidth]{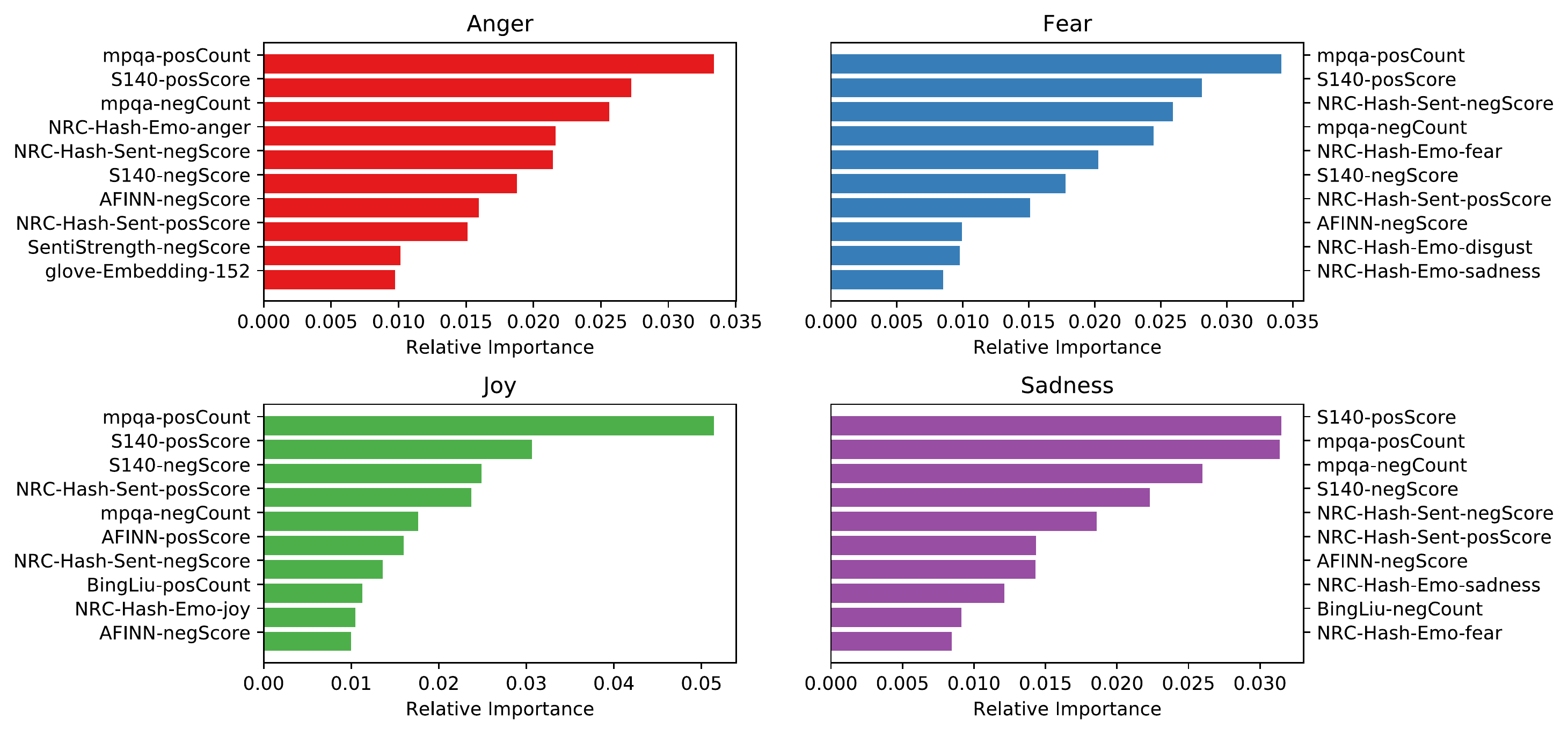}
	\caption{Relative Feature Importance of Various Emotions}
	\label{fig:feature_importance}
\end{figure*}

The relative feature importance can be assessed by the relative depth of the feature used as a decision node in the tree. Features used at the top of the tree contribute to the final prediction decision of a larger fraction of the input samples. The expected fraction of the samples they contribute to can thus be used as an estimate of the relative importance of the features. By averaging the measure over several randomized trees, the variance of the estimate can be reduced and used as a measure of relative feature importance.
In Figure \ref{fig:feature_importance} feature importance graphs are plotted for each emotion to infer which features are playing the major role in identifying emotional intensity in tweets. +/-EffectWordNet \cite{choi2014+}, NRC Hashtag Sentiment Lexicon, Sentiment140 Lexicon \cite{mohammad2013nrc} and NRC Hashtag Emotion Lexicon \cite{mohammad2015using} are playing the most important role. 

\subsection{System Limitations}
It is important to understand how the model performs in different scenarios. Table \ref{examples} analyzes when the system performs the best and worst for each emotion.
Since the features used are mostly lexicon based, the system has difficulties in capturing the overall sentiment and it leads to amplifying or vanishing intensity signals. For instance,
in example 4 of fear \textbf{louder} and \textbf{shaking} lexicons imply fear but overall sentence doesn't imply fear. A similar pattern can be found in the \nth{4} example of Anger and \nth{3} example of Joy. The system has difficulties in understanding of sarcastic tweets, for instance, in the \nth{3} tweet of Anger the user expressed anger but used \textbf{lol} which is used in a positive sense most of the times and hence the system did a bad job at predicting intensity. The system also fails in predicting sentences having deeper emotion and sentiment which humans can understand with a little context. For example, in sample 4 of sadness, the tweet refers to post travel blues which humans can understand. But with little context, it is difficult for the system to accurately estimate the intensity. The performance is poor with very short sentences as there are fewer indicators to provide a reasonable estimate.

\begin{table*}
\begin{tabular}{|c|c|c|c|}\hline
\textbf{Emotion} & \textbf{Tweet} & \textbf{Gold Int.} & \textbf{Pred. Int.} \\ \hline \hline
\multirow{9}{*}{\textbf{Anger}} & \multicolumn{1}{m{10cm}|}{@Claymakerbigsi @toghar11 @scott\_mulligan\_ @BoxingFanatic\_ Fucker blocked me 2 years ago over a question lol \- proper holds a grudge old Joe} & 0.625 & 0.6245 \\\cline{2-4}
& \multicolumn{1}{m{10cm}|}{We are raging angry.=1/2 bil \$ for 2 pro Liars.(Actors) the most useless people in america Where is ours for working 100 X harder? @FoxNews} & 0.667 & 0.6665 \\\cline{2-4}
& \multicolumn{1}{m{10cm}|}{dammit @TMobile whays going on!!! \includegraphics[height=1em]{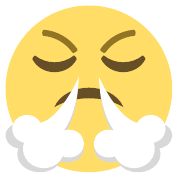}\includegraphics[height=1em]{1f624.pdf}\includegraphics[height=1em]{1f624.pdf}\includegraphics[height=1em]{1f624.pdf} lol  \#smh \#mobilefails} & 0.792 & 0.4062 \\\cline{2-4} 
& \multicolumn{1}{m{10cm}|}{People are \#hurt and \#angry and it's hard to know what to do with that \#anger Remember, at the end of the day, we're all \#humans \#bekind} & 0.250 & 0.6040 \\ \hline \hline

\multirow{5}{*}{\textbf{Fear}}    & \multicolumn{1}{m{10cm}|}{Onus is on \#Pak to act against \#terror groups which find safe havens and all types of support for cross border terror: \#MEA} & 0.667 & 0.6673 \\ \cline{2-4} 
& \multicolumn{1}{m{10cm}|}{Ffs dreadful defending} & 0.479 & 0.4795 \\ \cline{2-4}
& \multicolumn{1}{m{10cm}|}{\includegraphics[height=1em]{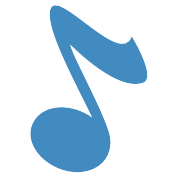}  OLD FISH} & 0.070 & 0.5028 \\ \cline{2-4} 
& \multicolumn{1}{m{10cm}|}{@MannersAboveAll *laughs louder this time, shaking my head* That was really cheesy, wasn't it?}  & 0.083  & 0.4936 \\ \hline \hline

\multirow{8}{*}{\textbf{Joy}} & \multicolumn{1}{m{10cm}|}{@headfirst\_dom I often imagine hoe our moon would feel meeting the jovial moons which are all special} & 0.500 & 0.5002 \\\cline{2-4}
& \multicolumn{1}{m{10cm}|}{Your attitude toward your struggles is equally as important as your actions to work through them.} & 0.340 & 0.3397 \\\cline{2-4}
& \multicolumn{1}{m{10cm}|}{Oi @THEWIGGYMESS you've absolutely fucking killed me.. 30 mins later im still crying with laughter.. Grindah.. Grindah... \includegraphics[height=1em]{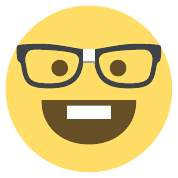}  hahahahahahaha} & 0.847 & 0.3726 \\\cline{2-4}
& \multicolumn{1}{m{10cm}|}{@WuffinArts :c You have my most heartfelt condolences. I'm glad it passed with levity and love in it's heart.} & 0.188  & 0.5872 \\ \hline \hline
\multirow{5}{*}{\textbf{Sadness}} & \multicolumn{1}{m{10cm}|}{@nytimes media celebrated Don King endorsing \#Obama in 08 and 12 now criticize him for endorsing \#Trump who wants new Civil Rights era\- sad}  & 0.562 & 0.5623 \\\cline{2-4} 
& \multicolumn{1}{m{10cm}|}{@AFCGraMaChroi oh, sorry if I've discouraged you \includegraphics[height=1em]{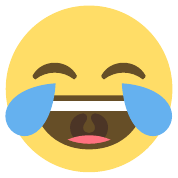}} & 0.340 & 0.3397 \\\cline{2-4} 
& \multicolumn{1}{m{10cm}|}{oh, btw - after a 6 month depression-free time I got a relapse now... superb \#depression} & 0.917 & 0.462 \\\cline{2-4} 
& \multicolumn{1}{m{10cm}|}{Ibiza blues hitting me hard already wow} & 0.833 & 0.4247              \\ \hline 
\end{tabular}
\caption{\label{examples} Sample tweets where our system's prediction is best and worst.}
\end{table*}

\section{Future Work \& Conclusion}
The paper studies the effectiveness of various affect lexicons word embeddings to estimate emotional intensity in tweets. A light-weight easy to use affect computing framework (EmoInt) to facilitate ease of experimenting with various lexicon features for text tasks is open-sourced. It provides plug and play access to various feature extractors and handy scripts for creating ensembles. 

Few problems explained in the analysis section can be resolved with the help of sentence embeddings which take the context information into consideration. The features used in the system are generic enough to use them in other affective computing tasks on social media text, not just tweet data. Another interesting feature of lexicon-based systems
is their good run-time performance during prediction, future work to benchmark the performance of the system can prove vital for deploying in a real-world setting.

\section*{Acknowledgement}
We would like to thank the organizers of the WASSA-2017 Shared Task on Emotion Intensity, for providing the data, the guidelines and timely support.


\bibliography{emnlp2017}

\begin{thebibliography}{}
\expandafter\ifx\csname natexlab\endcsname\relax\def\natexlab#1{#1}\fi

\bibitem[{Baccianella et~al.(2010)Baccianella, Esuli, and
  Sebastiani}]{baccianella2010sentiwordnet}
Stefano Baccianella, Andrea Esuli, and Fabrizio Sebastiani. 2010.
\newblock Sentiwordnet 3.0: An enhanced lexical resource for sentiment analysis
  and opinion mining.
\newblock In {\em LREC\/}. volume~10, pages 2200--2204.

\bibitem[{Bravo-Marquez et~al.(2016)Bravo-Marquez, Frank, Mohammad, and
  Pfahringer}]{bravo2016determining}
Felipe Bravo-Marquez, Eibe Frank, Saif~M Mohammad, and Bernhard Pfahringer.
  2016.
\newblock Determining word--emotion associations from tweets by multi-label
  classification.
\newblock In {\em WI'16\/}. IEEE Computer Society, pages 536--539.

\bibitem[{Bravo-Marquez et~al.(2015)Bravo-Marquez, Frank, and
  Pfahringer}]{bravo2015unlabelled}
Felipe Bravo-Marquez, Eibe Frank, and Bernhard Pfahringer. 2015.
\newblock From unlabelled tweets to twitter-specific opinion words.
\newblock In {\em Proceedings of the 38th International ACM SIGIR Conference on
  Research and Development in Information Retrieval\/}. ACM, pages 743--746.

\bibitem[{Brown et~al.(1992)Brown, Desouza, Mercer, Pietra, and
  Lai}]{brown1992class}
Peter~F Brown, Peter~V Desouza, Robert~L Mercer, Vincent J~Della Pietra, and
  Jenifer~C Lai. 1992.
\newblock Class-based n-gram models of natural language.
\newblock {\em Computational linguistics\/} 18(4):467--479.

\bibitem[{Chen and Guestrin(2016)}]{chen2016xgboost}
Tianqi Chen and Carlos Guestrin. 2016.
\newblock Xgboost: A scalable tree boosting system.
\newblock In {\em Proceedings of the 22Nd ACM SIGKDD International Conference
  on Knowledge Discovery and Data Mining\/}. ACM, pages 785--794.

\bibitem[{Choi and Wiebe(2014)}]{choi2014+}
Yoonjung Choi and Janyce Wiebe. 2014.
\newblock +/-effectwordnet: Sense-level lexicon acquisition for opinion
  inference.
\newblock In {\em EMNLP\/}. pages 1181--1191.

\bibitem[{Eisner et~al.(2016)Eisner, Rockt{\"a}schel, Augenstein,
  Bo{\v{s}}njak, and Riedel}]{eisner2016emoji2vec}
Ben Eisner, Tim Rockt{\"a}schel, Isabelle Augenstein, Matko Bo{\v{s}}njak, and
  Sebastian Riedel. 2016.
\newblock emoji2vec: Learning emoji representations from their description.
\newblock {\em arXiv preprint arXiv:1609.08359\/} .

\bibitem[{Hu and Liu(2004)}]{hu2004mining}
Minqing Hu and Bing Liu. 2004.
\newblock Mining and summarizing customer reviews.
\newblock In {\em Proceedings of the tenth ACM SIGKDD international conference
  on Knowledge discovery and data mining\/}. ACM, pages 168--177.

\bibitem[{Mikolov et~al.(2013)Mikolov, Sutskever, Chen, Corrado, and
  Dean}]{mikolov2013distributed}
Tomas Mikolov, Ilya Sutskever, Kai Chen, Greg~S Corrado, and Jeff Dean. 2013.
\newblock Distributed representations of words and phrases and their
  compositionality.
\newblock In {\em Advances in neural information processing systems\/}. pages
  3111--3119.

\bibitem[{Mohammad(2017)}]{mohammad2017word}
Saif~M Mohammad. 2017.
\newblock Word affect intensities.
\newblock {\em arXiv preprint arXiv:1704.08798\/} .

\bibitem[{Mohammad and Bravo-Marquez(2017{\natexlab{a}})}]{affectivetweets}
Saif~M. Mohammad and Felipe Bravo-Marquez. 2017{\natexlab{a}}.
\newblock Emotion intensities in tweets .

\bibitem[{Mohammad and Bravo-Marquez(2017{\natexlab{b}})}]{MohammadB17starsem}
Saif~M. Mohammad and Felipe Bravo-Marquez. 2017{\natexlab{b}}.
\newblock Emotion intensities in tweets.
\newblock In {\em Proceedings of the sixth joint conference on lexical and
  computational semantics (*Sem)\/}. Vancouver, Canada.

\bibitem[{Mohammad and Bravo-Marquez(2017{\natexlab{c}})}]{wassatask2017}
Saif~M. Mohammad and Felipe Bravo-Marquez. 2017{\natexlab{c}}.
\newblock Wassa-2017 shared task on emotion intensity.
\newblock EMNLP 2017 Workshop on Computational Approaches to Subjectivity,
  Sentiment, and Social Media (WASSA), Copenhagen, Denmark.

\bibitem[{Mohammad and Kiritchenko(2015)}]{mohammad2015using}
Saif~M Mohammad and Svetlana Kiritchenko. 2015.
\newblock Using hashtags to capture fine emotion categories from tweets.
\newblock {\em Computational Intelligence\/} 31(2):301--326.

\bibitem[{Mohammad et~al.(2013)Mohammad, Kiritchenko, and
  Zhu}]{mohammad2013nrc}
Saif~M Mohammad, Svetlana Kiritchenko, and Xiaodan Zhu. 2013.
\newblock Nrc-canada: Building the state-of-the-art in sentiment analysis of
  tweets.
\newblock {\em arXiv preprint arXiv:1308.6242\/} .

\bibitem[{Mohammad and Turney(2010)}]{mohammad2010emotions}
Saif~M Mohammad and Peter~D Turney. 2010.
\newblock Emotions evoked by common words and phrases: Using mechanical turk to
  create an emotion lexicon.
\newblock In {\em Proceedings of the NAACL HLT 2010 workshop on computational
  approaches to analysis and generation of emotion in text\/}. Association for
  Computational Linguistics, pages 26--34.

\bibitem[{Nielsen(2011)}]{nielsen2011new}
Finn~{\AA}rup Nielsen. 2011.
\newblock A new anew: Evaluation of a word list for sentiment analysis in
  microblogs.
\newblock {\em arXiv preprint arXiv:1103.2903\/} .

\bibitem[{Owoputi et~al.(2013)Owoputi, O'Connor, Dyer, Gimpel, Schneider, and
  Smith}]{owoputi2013improved}
Olutobi Owoputi, Brendan O'Connor, Chris Dyer, Kevin Gimpel, Nathan Schneider,
  and Noah~A Smith. 2013.
\newblock Improved part-of-speech tagging for online conversational text with
  word clusters.
\newblock Association for Computational Linguistics.

\bibitem[{Pedregosa et~al.(2011)Pedregosa, Varoquaux, Gramfort, Michel,
  Thirion, Grisel, Blondel, Prettenhofer, Weiss, Dubourg, Vanderplas, Passos,
  Cournapeau, Brucher, Perrot, and Duchesnay}]{scikit-learn}
F.~Pedregosa, G.~Varoquaux, A.~Gramfort, V.~Michel, B.~Thirion, O.~Grisel,
  M.~Blondel, P.~Prettenhofer, R.~Weiss, V.~Dubourg, J.~Vanderplas, A.~Passos,
  D.~Cournapeau, M.~Brucher, M.~Perrot, and E.~Duchesnay. 2011.
\newblock Scikit-learn: Machine learning in {P}ython.
\newblock {\em Journal of Machine Learning Research\/} 12:2825--2830.

\bibitem[{Pennington et~al.(2014)Pennington, Socher, and
  Manning}]{pennington2014glove}
Jeffrey Pennington, Richard Socher, and Christopher~D Manning. 2014.
\newblock Glove: Global vectors for word representation.
\newblock In {\em EMNLP\/}. volume~14, pages 1532--1543.

\bibitem[{Petrovic et~al.(2010)Petrovic, Osborne, and
  Lavrenko}]{petrovic2010edinburgh}
Sasa Petrovic, Miles Osborne, and Victor Lavrenko. 2010.
\newblock The edinburgh twitter corpus.
\newblock In {\em Proceedings of the NAACL HLT 2010 Workshop on Computational
  Linguistics in a World of Social Media\/}. pages 25--26.

\bibitem[{Thelwall et~al.(2010)Thelwall, Buckley, Paltoglou, Cai, and
  Kappas}]{thelwall2010sentiment}
Mike Thelwall, Kevan Buckley, Georgios Paltoglou, Di~Cai, and Arvid Kappas.
  2010.
\newblock Sentiment strength detection in short informal text.
\newblock {\em Journal of the American Society for Information Science and
  Technology\/} 61(12):2544--2558.

\end{thebibliography}
\bibliographystyle{emnlp_natbib}

\end{document}